\documentclass[MSNbibl,seceqn,nameyear,dvips]{arxbj}
\usepackage{graphicx}

\aid{0}
\volume{19}
\issue{4}
\pubyear{2013}
\firstpage{1378}
\lastpage{1390}
\doi{10.3150/12-BEJSP17} 

\makeatletter
\newcommand{\rrVert}{\Vert}
\newcommand{\llVert}{\Vert}

\newtheorem{theorem}{Theorem}[section]

\newcommand{\R}{{\mathbb R}}
\newcommand{\bx}{\mathbf{x}}
\newcommand{\ba}{\mathbf{a}}
\newcommand{\bb}{\mathbf{b}}
\newcommand{\by}{\mathbf{y}}
\newcommand{\bz}{\mathbf{z}}
\newcommand{\bs}{\mathcal{S}}
\newcommand{\bc}{\mathcal{C}}
\newcommand{\G}{g}

\makeatother

\begin{document}
\begin{frontmatter}

\title{On statistics, computation and scalability}
\runtitle{On statistics, computation and scalability}

\begin{aug}
\author{\fnms{Michael I.} \snm{Jordan}\corref{}\ead[label=e1]{jordan@stat.berkeley.edu}\ead[label=e2,url]{www.cs.berkeley.edu/\textasciitilde jordan}}
\runauthor{M.I. Jordan} 
\address{Department of Statistics and Department of EECS,
University of California, Berkeley, CA, USA. \printead{e1}; \printead{e2}}
\end{aug}


%
\begin{abstract}
How should statistical procedures be designed so as to be scalable
computationally to the massive datasets that are increasingly the
norm? When coupled with the requirement that an answer to an inferential
question be delivered within a certain time budget, this question has
significant repercussions for the field of statistics. With the goal of
identifying ``time-data tradeoffs,'' we investigate some of the statistical
consequences of computational perspectives on scability, in particular
divide-and-conquer methodology and hierarchies of convex relaxations.
\end{abstract}


\end{frontmatter}

The fields of computer science and statistics have undergone mostly separate
evolutions during their respective histories. This is changing, due in
part to
the phenomenon of ``Big Data.'' Indeed, science and technology are currently
generating very large datasets and the gatherers of these data have increasingly
ambitious inferential goals, trends which point towards a future in which
statistics will be forced to deal with problems of scale in order to remain
relevant. Currently the field seems little prepared to meet this challenge.
To the key question ``Can you guarantee a certain level of inferential accuracy
within a certain time budget even as the data grow in size?'' the field is
generally silent. Many statistical procedures either have unknown runtimes
or runtimes that render the procedure unusable on large-scale data. Although
the field of sequential analysis provides tools to assess risk after a
certain number of data points have arrived, this is different from an
algorithmic analysis that predicts a relationship between time and risk.
Faced with this situation, gatherers of large-scale data are often
forced to
turn to ad hoc procedures that perhaps do provide algorithmic
guarantees but
which may provide no statistical guarantees and which in fact may have poor
or even disastrous statistical properties.

On the other hand, the field of computer science is also currently poorly
equipped to provide solutions to the inferential problems associated
with Big Data. Database researchers rarely view the data in a database as
noisy measurements on an underlying population about which inferential
statements are desired. Theoretical computer scientists are able to provide
analyses of the resource requirements of algorithms (e.g., time and space),
and are often able to provide comparative analyses of different algorithms
for solving a given problem, but these problems rarely refer to inferential
goals. In particular, the notion that it may be possible to save on computation
because of the growth of statistical power as problem instances grow in size
is not (yet) a common perspective in computer science.

In this paper we discuss some recent research initiatives that aim to draw
computer science and statistics closer together, with particular
reference to
``Big Data'' problems. There are two main underlying perspectives driving
these initiatives, both of which present interesting conceptual challenges
for statistics. The first is that large computational problems are
often usefully addressed via some notion of ``divide-and-conquer.''
That is,
the large problem is divided into subproblems that are hopefully
simpler than
the original problem, these subproblems are solved (sometimes again
with a
divide-and-conquer strategy) and the solutions are pieced together to solve
the original problem. In the statistical setting, one natural subdivision
strategy involves breaking the data into subsets. The estimator of
interest is
applied to the subsets and the results are combined. The challenge in the
statistical setting is that the analysis of subsets of data may present
different statistical properties than the overall dataset. For example,
confidence intervals based on subsets of data will generally be wider than
confidence intervals based on the original data; thus, care must be taken
that the overall divide-and-conquer procedure yields a correctly calibrated
interval.

The second perspective involves a notion of ``algorithmic weakening,'' whereby
we do not consider a single algorithm for solving an inference problem, but
instead consider a hierarchy of algorithms that are ordered by computational
complexity. As data accrue, we want to back off to cheaper algorithms that
run more quickly and deliver a result that would be viewed as being of poorer
quality from a classical algorithmic point of view. We hope to do this in
a way such that the increasing statistical strength of the data compensate
for the poor algorithmic quality, so that in fact the overall quality of
inference increases as data accrue, even if we impose a computational budget.
The challenge is to do this in a theoretically sound way.

The remainder of the paper is organized into three subsections, the first
two concerned with divide-and-conquer algorithms, and the third concerned
with algorithmic weakening.

\section{Bag of little bootstraps}

In this section we consider the core inferential problem of evaluating
the quality of point estimators, a problem that is addressed by the
bootstrap [\citet{Efron}] and related resampling-based methods. The material
in this section summarizes research described in \citet{Kleiner}.

The usual implementation of the bootstrap involves the
``computationally-intensive''
procedure of resampling the original data with replacement, applying
the estimator
to each such bootstrap resample, and using the resulting distribution
as an
approximation to the true sampling distribution and thereby computing
(say) a confidence interval. A~notable virtue of this approach in the
setting of modern distributed computing platforms is that it readily
parallelizes -- each bootstrap resample can be processed independently
by the
processors of a ``cloud computer.'' Thus in principle it should be possible
to compute bootstrap confidence intervals in essentially the same
runtime as is
required to compute the point estimate. In the massive data setting, however,
there is a serious problem: each bootstrap resample is itself massive (roughly
0.632 times the original dataset size). Processing such resampled
datasets can
overwhelm available computational resources; for example, with a
terabyte of
data it may not be straightforward to send a few hundred resampled datasets,
each of size 632 gigabytes, to a set of distributed processors on a network.

An appealing alternative is to work with subsets of data, an instance
of the
divide-and-conquer paradigm. Existing examples of this alternative
approach include
subsampling [\citet{Politis-etal}] and the $m$-out-of-$n$
bootstrap [\citet{Bickel}].
In both cases, the idea is that a dataset of size $n$ can be processed into
multiple sets of size $m$ (there are ${n \choose m}$ such subsets in
the case
of sampling without replacement), and the estimator computed on each set.
This yields fluctuations in the values of the point estimate. The challenge
is the one referred to earlier -- these fluctuations are on the wrong scale,
being based on datasets that are smaller than the original dataset.
Both subsampling
and the $m$-out-of-$n$ bootstrap assume that an analytical correction factor
is available (e.g., $\sqrt{m/n}$) to rescale the confidence intervals
obtained from the sampled subsets. This renders these procedures
somewhat less
``user-friendly'' than the bootstrap, which requires no such correction factor.

There are situations in which the bootstrap is known to be inconsistent,
and where subsampling and the $m$-out-of-$n$ bootstrap are consistent;
indeed, the search for a broader range of consistency results was the original
motivation for exploring these methods. On the other hand, finite sample
results do not necessarily favor the consistent procedures over the
bootstrap [see, e.g., \citet{Samworth}]. The intuition is as follows.
For small values of $m$, the procedure performs poorly, because each
estimate is highly noisy. As $m$ increases, the noise decreases and
performance improves. For large values of $m$, however, there are too
few subsamples, and performance again declines. In general it is difficult
to find the appropriate value of $m$ for a given problem.

In recent work, \citet{Kleiner} have explored a new procedure, the
``Bag of Little
Bootstraps'' (BLB), which targets computational efficiency, but which also
alleviates some of the difficulties of subsampling, the $m$-out-of-$n$
bootstrap and the bootstrap, essentially by combining aspects of these
procedures. The basic idea of BLB is as follows. Consider a subsample
of size $m$ (taken either with replacement or without replacement).
Note that this subsample is itself a random sample from the population,
and thus the empirical distribution formed from this subsample is an
approximation to the population distribution. It is thus reasonable
to sample from this empirical distribution as a plug-in proxy for
the population. In particular, there is nothing preventing us from
sampling $n$ times from this empirical distribution (rather than $m$
times). That is, we can implement the bootstrap on the correct scale
using this subsample, simply by using it to generate multiple bootstrap
samples of size $n$. Now, the resulting confidence interval will be
a bona fide bootstrap confidence interval, but it will be noisy, because
it is based on a (small) subsample. But we can proceed as in subsampling,
repeating the procedure multiple times with randomly chosen subsamples.
We obtain a set of bootstrap confidence intervals, which we combine
(e.g., by averaging) to yield the overall bootstrap confidence interval.

%
\begin{figure}

\includegraphics{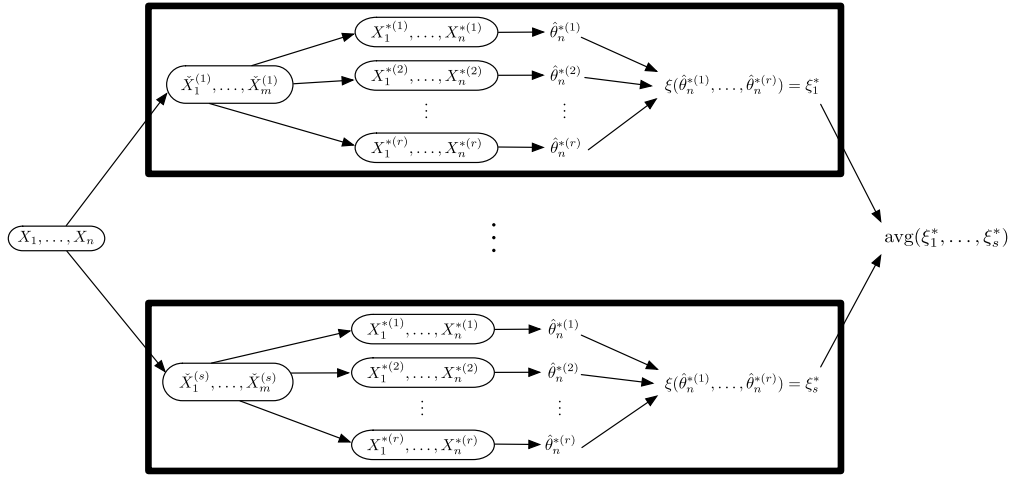}

\caption{The BLB procedure. From the original dataset, $\{X_1,\ldots,
X_n\}$, $s$ subsamples of size $m$ are formed. From each of these
subsamples, $r$ bootstrap resamples are formed, each of which are
conceptually of size~$n$ (but would generally be stored as weighted
samples of size $m$). The resulting bootstrap estimates of risk are
averaged. In a parallel implementation of BLB, the boxes in the
diagram would correspond to separate processors; moreover, the
bootstrap resampling within a box could also be parallelized.}
\label{figblb}
\end{figure}

The procedure is summarized in Figure~\ref{figblb}. We see that
BLB is composed of two nested procedures, with the inner procedure
being the bootstrap applied to a subsample, and the outer procedure
being the combining of these multiple bootstrap estimates. From a
computational point of view, the BLB procedure can be mapped onto a
distributed computing architecture by letting each subsample be processed
by a separate processor. Note that BLB has the virtue that the subsamples
sent to each processors are small (of~size~$m$). Moreover, although
the inner loop of bootstrapping conceptually creates multiple resampled
datasets of size $n$, it is not generally necessary to create actual
datasets of size $n$; instead we form weighted datasets of size $m$.
(Also, the weights can be obtained as draws from a Poisson distribution
rather than via explicit multinomial sampling.) Such is the case, for
example, for estimators that are plug-in functionals of the empirical
distribution.

\begin{figure}

\includegraphics{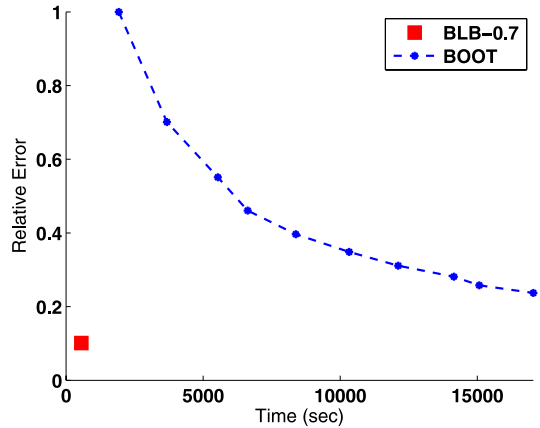}

\caption{Univariate confidence intervals for a logistic regression
on 6\,000\,000 data points using a 80-core distributed computing platform.
The data were generated synthetically and thus the ground-truth sampling
distribution was available via Monte Carlo; the $y$-axis is a measure
of relative error with respect to this ground truth. See \citet{Kleiner}
for further details.}
\label{figblb-results}
\end{figure}

An example taken from \citet{Kleiner} serves to illustrate the very
substantial computational gains that can be reaped from this approach.
Consider computing bootstrap confidence intervals for the estimates
of the individual components of the parameter vector in logistic
regression, where the covariate vector has dimension 3000 and there
are 6\,000\,000 data points, and where a distributed computing platform
involving 80 cores (8 cores on each of ten processors) is available.
To implement the bootstrap at this scale, we can parallelize the
logistic regression, and sequentially process the bootstrap resamples.
Results from carrying out such a procedure are shown as the dashed
curve in Figure~\ref{figblb-results}, where we see that the processing
of each bootstrap resample requires approximately 2000 seconds of
processing time.
%
The other natural approach is to implement a parallel version of BLB
as we have discussed, where each processor executes the bootstrap on
$m(n) = n^\gamma$ points via weighted logistic regressions. The results
are also shown in Figure~\ref{figblb-results}, as a single dot in the
lower-left corner of the figure. Here $\gamma$ is equal to $0.7$.
We see that BLB has finished in less time than is required for a single
iteration of the bootstrap on the full dataset, and indeed in less than
750 seconds has delivered an accuracy that is significantly better than
that obtained by the bootstrap after 15\,000 seconds.

Thus we see that there is a very strong synergy between a particular way
to organize bootstrap-style computation and the capabilities of modern
distributed computing platforms. Moreover, although the development of
BLB was motivated by the computational imperative, it can be viewed as
a novel statistical procedure to be compared to the bootstrap and
subsampling according to more classical criteria; indeed, \citet{Kleiner}
present experiments that show that even on a single processor BLB converges
faster than the bootstrap and it is less sensitive to the choice of $m$
than subsampling and the $m$-out-of-$n$ bootstrap.

There is much more to be done along these lines. For example, stratified
sampling and other sophisticated sampling schemes can likely be mapped
in useful ways to distributed platforms. For dependent data, one wants to
resample in ways that respect the dependence, and this presumably favors
certain kinds of data layout and algorithms over others. In general, for
statistical inference to not run aground on massive datasets, we need
for statistical thinking to embrace computational thinking.

\section{Divide-and-conquer matrix factorization}

Statistics has long exploited matrix analysis as a core computational tool,
with linear models, contingency tables and multivariate analysis providing
well-known examples. Matrices continue to be the focus of much recent
computationally-focused research, notably as representations of graphs
and networks and in collaborative filtering applications. At the core
of a significant number of analysis procedures is the notion of
\emph{matrix factorization}, with the singular value decomposition
(SVD) providing a canonical example. The SVD in particular yields
low-rank representations of matrices, which often maps directly to
modeling assumptions.

Many matrix factorization procedures, including the SVD, have cubic
algorithmic complexity in the row or column dimension of the matrix.
This is overly burdensome in many applications, in statistics and beyond,
and there is a strong motivation for applied linear algebra researchers
to devise efficient ways to exploit parallel and distributed hardware
in the computation of the SVD and other factorizations. One could
take the point of view that this line of research is outside of the purview
of statistics; that statisticians should simply keep an eye on developments.
But this neglects the fact that as problems grow in size, it is the particular
set of modeling assumptions at play that determine whether efficient
algorithms are available, and in particular whether computationally-efficient
approximations can be obtained that are statistically meaningful.

As a particularly salient example, in many statistical applications
involving large-scale matrices it is often the case that many entries
of a matrix are missing. Indeed, the quadratic growth in the number of
entries of a matrix is often accompanied by a linear growth in the rate
of observation of matrix entries, such that at large scale the vast majority
of matrix entries are missing. This is true in many large-scale collaborative
filtering applications, where, for example, most individuals will have
rated a small fraction of the overall set of items (e.g., books or movies).
It is also true in many graph or network analysis problems, where each node
is connected to a vanishingly small fraction of the other nodes.

\begin{figure}

\includegraphics{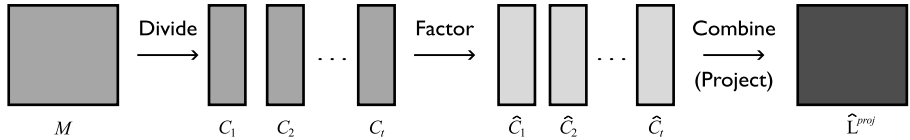}

\caption{The DFC pipeline. The matrix $M$ is partitioned according to its
columns and the resulting submatrices $\{C_i\}$ are factored in parallel.
The factored forms, $\{\hat{C}_i\}$, are then transmitted to a central
location where they are combined into an overall factorization $\hat
{L}^{\mathrm{proj}}$.}
\label{figdfc-pipeline}
\end{figure}

In recent work, \citet{Mackey} have studied a divide-and-conquer methodology
for matrix factorization that aims to exploit parallel hardware platforms.
Their framework, referred to as \emph{Divide-Factor-Combine} (DFC),
is rather simple from an algorithmic point of view -- a matrix is
partitioned according to its columns or rows, matrix factorizations
are obtained (in parallel) for each of the submatrices using a
``base algorithm'' that is one of the standard matrix factorization
methods, and the factorizations are combined to obtain an overall factorization
(see Figure~\ref{figdfc-pipeline}).
%
The question is how to design such a pipeline so as to retain the statistical
guarantees of the base algorithm, while providing computational speed-ups.

Let us take the example of \emph{noisy matrix completion} [see,
e.g., \citet{CandesPlan2010}], where we model a matrix $M \in\mathbb
{R}^{m \times n}$
as the sum of a low-rank matrix $L_0$ (with rank $r \ll\min(m,n)$)
and a
noise matrix $Z$:
\[
M = L_0 + Z,
\]
and where only a small subset of the entries of $M$ are observed. Letting
$\Omega$ denote the set of indices of the observed entries, the goal
is to
estimate $L_0$ given $\{M_{ij}\dvtx  (i,j) \in\Omega\}$. This goal can be
formulated in terms of an optimization problem:
%
\begin{eqnarray}
&\displaystyle\min_L \qquad\operatorname{rank}(L)&
\nonumber\\[-8pt]\\[-8pt]
&\mbox{subject to } \displaystyle \sum_{(i,j) \in\Omega} (L_{ij} -
M_{ij})^2 \leq \Delta^2,&\nonumber
\end{eqnarray}
for a specified value $\Delta$. This problem is computationally intractable,
so it is natural to consider replacing the rank function with its tightest
convex relaxation, the nuclear norm, yielding the following convex
optimization problem [\citet{CandesPlan2010}]:
%
\begin{eqnarray}\label{eqnuclear-norm}
&\displaystyle\min_L \qquad\|{L}\|_*&
\nonumber\\[-8pt]\\[-8pt]
&\mbox{subject to } \displaystyle \sum_{(i,j) \in\Omega} (L_{ij} -
M_{ij})^2 \leq \Delta^2,&\nonumber
\end{eqnarray}
where $\|{L}\|_*$ denotes the nuclear norm of $L$ (the sum of the singular
values of $L$).

\citet{CandesPlan2010} have provided conditions under which the
solution to
equation (\ref{eqnuclear-norm}) recovers the matrix $L_0$ despite the
potentially
large number
of unobserved entries of~$M$. These conditions involve a structural
assumption on the matrix $L_0$ (that its singular vectors should not be
too sparse or too correlated) and a sampling assumption for~$\Omega$
(that the entries of the matrix are sampled uniformly at random).
We will refer to the former assumption as ``$(\mu,r)$-coherence,''
referring to \citet{CandesPlan2010} for technical details.
Building on work by \citet{Recht}, \citet{Mackey} have proved
the following theorem, in the spirit of \citet{CandesPlan2010}
but with weaker conditions:
%
\begin{theorem}
\label{thmconvex-mc-noise}
Suppose that $L_0$ is $(\mu,r)$-coherent and $s$ entries of $M$ are
observed at locations $\Omega$ sampled uniformly without replacement,
where
\[
s \geq32 \mu r(m+n) \log^2(m+n).
\]
Then, if $\sum_{(i,j) \in\Omega} (M_{ij} - L_{0,ij})^2 \leq\Delta
^2$ a.s.,
the minimizer $\hat{L}$ of equation (\ref{eqnuclear-norm}) satisfies
\[
\|{L_0 - \hat{L}}\|_F \leq c_e \sqrt{mn}
\Delta,
\]
with high probability, where $c_e$ is a universal constant.
\end{theorem}
Note in particular that the required sampling rate $s$ is a vanishing fraction
of the total number of entries of $M$.

Theorem~\ref{thmconvex-mc-noise} exemplifies the kind of theoretical guarantee
that one would like to retain under the DFC framework. Let us therefore consider
a particular example of the DFC framework, referred to as ``\textsc{DFC-Proj}''
by \citet{Mackey}, in which the ``divide'' step consists in the partitioning
of the columns of $M$ into $t$ submatrices each having $l$ columns
(assuming for simplicity that $l$ divides $n$), the ``factor'' step
involves solving the nuclear norm minimization problem in equation
(\ref{eqnuclear-norm})
for each submatrix (in parallel), and the ``combine'' step consists of a
projection step in which the $t$ low-rank approximations are projected
onto a
common subspace.\footnote{In particular, letting
$\{\hat{C}_1, \hat{C}_2,\ldots, \hat{C}_t\}$ denote
the $t$ low-rank approximations, we can project these submatrices
onto the column space of any one of these submatrices, for example,
$\hat{C}_1$. Mackey, Talwalkar and Jordan
(\citeyear{Mackey}) also propose an ``ensemble'' version of this procedure
in which
the low-rank submatrices are projected onto each other (i.e., onto
$\hat{C}_k$,
for $k = 1,\ldots, t$) and the resulting projections are averaged.}
Retaining a theoretical guarantee for \textsc{DFC-Proj} from that of
its base algorithm essentially involves ensuring that the
$(\mu,r)$-coherence of the overall matrix is not increased very much
in the
random selection of submatrices in the ``divide'' step, and that the
low-rank approximations obtained in the ``factor'' step are not far
from the low-rank approximation that would be obtained from the overall
matrix.

In particular, \citet{Mackey} establish the following theorem:\footnote{We
have simplified the statement of the theorem to streamline the presentation;
see Mackey, Talwalkar and Jordan (\citeyear{Mackey}) for the full result.}
%
\begin{theorem}
Suppose that $L_0$ is $(\mu,r)$-coherent and that $s$ entries of $M$ are
observed at locations $\Omega$ sampled uniformly without replacement.
Then, if $\sum_{(i,j) \in\Omega} (M_{ij} - L_{0,ij})^2 \leq\Delta
^2$ a.s.,
and the base algorithm in the ``factor'' step of \textsc{DFC-Proj} involves
solving the optimization problem of equation (\ref{eqnuclear-norm}),
it suffices to choose
\[
l \geq {c\mu^2 r^2(m+n)n \log^2(m+n)}/
\bigl(s\varepsilon^2\bigr)
\]
columns in the ``divide'' step to achieve
\[
\|{L_0 - \hat{L}}\|_F \leq(2+\varepsilon)
c_e \sqrt{mn} \Delta,
\]
with high probability.
\end{theorem}
Thus, the \textsc{DFC-Proj} algorithm achieves essentially the rate
established in
Theorem~\ref{thmconvex-mc-noise} for the nuclear norm minimization algorithm.
Moreover, if we set $s=\omega((m+n)\log^2(m+n))$, which is slightly
faster than
the lower bound in Theorem~\ref{thmconvex-mc-noise}, then we see that
$l/n \rightarrow0$.
That is, \textsc{DFC-Proj} succeeds even if only a vanishingly small
fraction of
the columns are sampled to form the submatrices in the ``divide'' step.

\begin{figure}
\begin{tabular}{@{}cc@{}}

\includegraphics{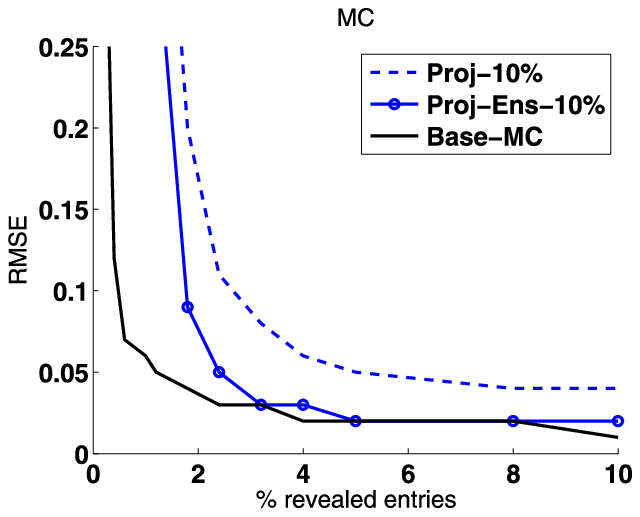}
 & \includegraphics{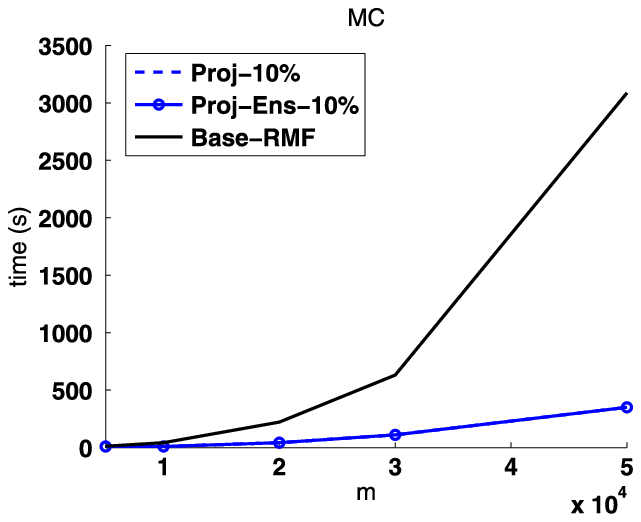}\\
(a) & (b)
\end{tabular}
\caption{Numerical experiments on matrix completion with DFC.
\textup{(a)} Accuracy for a state-of-the-art baseline algorithm
(``Base-MC'') and \textsc{DFC-Proj} (see Mackey, Talwalkar and Jordan
(\citeyear{Mackey}) for details),
as a function of the percentage of revealed entries.
\textup{(b)} Runtime as a function of matrix dimension
(these are square matrices, so $m = n$).}
\label{figdfc-proj-results}
\end{figure}

Figure~\ref{figdfc-proj-results} shows representative numerical results
in an experiment on matrix completion reported by \citet{Mackey}.
%
The leftmost figure shows that the accuracy (measured as root mean
square error) achieved by the ensemble version of \textsc{DFC-Proj}
is nearly the same as that of the base algorithm. The rightmost
figure shows that this accuracy is obtained at a fraction of the
computational cost required by the baseline algorithm.

\section{Convex relaxations}

The methods that we have discussed thus far provide a certain degree
of flexibility in the way inferential computations are mapped onto
a computing infrastructure, and this flexibility implicitly defines
a tradeoff between speed and accuracy. In the case of BLB the flexibility
inheres in the choice of $m$ (the subsample size) and in the case of DFC
it is the choice of $l$ (the submatrix dimension). In the work that
we discuss in this section the goal is to treat such tradeoffs explicitly.
To achieve this, \citet{Chandrasekaran} define a notion of ``algorithmic
weakening,'' in which a hierarchy of algorithms is ordered by both
computational efficiency and statistical efficiency. The problem
that they address is to develop a quantitative relationship among
three quantities: the number of data points, the runtime and the
statistical risk.

\citet{Chandrasekaran} focus on the denoising problem, an important
theoretical testbed in the study of high-dimensional
inference [cf. \citet{DonJ1998}]. The model is the following:
%
\begin{equation}\label{eqdenoise}
\by= \mathbf{x}^\ast+ \sigma\bz,
\end{equation}
where $\sigma> 0$, the noise vector $\bz\in\R^p$ is standard normal,
and the unknown parameter $\mathbf{x}^\ast$ belongs to a known subset
$\bs
\subset\R^p$.
The problem is to estimate $\mathbf{x}^\ast$ based on $n$ independent
observations
$\{\by_i\}_{i=1}^n$ of $\by$.

Consider a shrinkage estimator given by a projection of the sufficient
statistic $\bar{\by} = \frac{1}{n} \sum_{i=1}^n \by_i$ onto a
convex set
$\bc$ that is an outer approximation to $\bs$, that is, $\bs\subset\bc$:
%
\begin{equation}\label{eqshrink}
\hat{\bx}_n(\bc) = \arg\min_{\bx\in\R^p}\quad
\frac{1}{2}\llVert \bar{\by} - \bx\rrVert _{\ell_2}^2
\quad\mbox{s.t.}\quad \bx\in\bc.
\end{equation}
The procedure studied by \citet{Chandrasekaran} consists of a \emph{set}
of such projections, $\{\hat{\bx}_n(\bc_i)\}$, obtained from a hierarchy
of convex outer approximations,
\[
\bs\subseteq\cdots\subseteq\bc_3 \subseteq\bc_2
\subseteq\bc_1.
\]
The intuition is that for $i < j$, the estimator $\{\hat{\bx}_n(\bc
_i)\}$
will exhibit poorer statistical performance than $\{\hat{\bx}_n(\bc
_j)\}$,
given that $\bc_i$ is a looser approximation to $\bs$ than $\bc_j$,
but that
$\bc_i$ can be chosen to be a simpler geometrical object than $\bc
_j$, such
that it is computationally cheaper to optimize over $\bc_i$, and thus more
samples can be processed by the estimator $\{\hat{\bx}_n(\bc_i)\}$
in a given time frame, offsetting the increase in statistical risk.
Indeed, such \emph{convex relaxations} have been widely used to give
efficient approximation algorithms for intractable problems in computer
science [\citet{Vaz2004}], and much is known about the decrease in runtime
as one moves along the hierarchy of relaxations. To develop a time/data
tradeoff, what is needed is a connection to statistical risk as one moves
along the hierarchy.

\begin{figure}[b]

\includegraphics{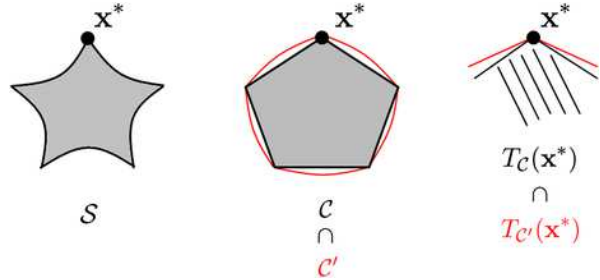}

\caption{(left) A signal set $\bs$ containing the
true signal $\mathbf{x}^\ast$; (middle) Two convex constraint sets
$\bc$ and~$\bc'$, where $\bc$ is the convex hull of $\bs$ and $\bc'$ is
a relaxation that is more efficiently computable than $\bc$; (right)
The tangent cone $T_\bc(\mathbf{x}^\ast)$ is contained inside the
tangent cone $T_{\bc'}(\mathbf{x}^\ast)$. Consequently,\vspace*{1pt} the Gaussian
squared-complexity $\G(T_\bc(\mathbf {x}^\ast) \cap B_{\ell_2}^p)$ is
smaller than the complexity $\G(T_{\bc'}(\mathbf{x}^\ast) \cap B_{\ell
_2}^p)$, so that the estimator $\hat{\bx}_n(\bc)$ requires fewer
samples than the estimator $\hat{\bx}_n(\bc')$ for a risk of at most
$1$.} \label{figrelax}
\end{figure}

\citet{Chandrasekaran} show that convex geometry provides such a connection.
Define the \emph{Gaussian squared-complexity} of a set $\mathcal{D}
\in\R^p$
as follows:
\[
\G(\mathcal{D}) = \mathbb{E} \Bigl[\sup_{\bolds{\delta} \in\mathcal{D}}  \langle\bolds {
\delta}, \bz\rangle^2 \Bigr],
\]
where the expectation is with respect to $\bz\sim\mathcal{N}(0,I_{p
\times p})$.
Given a closed convex set $\bc\in\R^p$ and a point $\ba\in\bc$,
define the
\emph{tangent cone} at $\ba$ with respect to $\bc$ as
%
\begin{equation}\label{eqtcone}
T_{\bc}(\ba) = \mathrm{cone}\{\bb-\ba\mid\bb\in\bc\},
\end{equation}
where $\mathrm{cone}(\cdot)$ refers to the conic hull of a set obtained
by taking nonnegative linear combinations of elements of the set.
(See Figure~\ref{figrelax} for a depiction of the geometry.)
%
Let\vspace*{1pt} $B_{\ell_2}^p$ denote the $\ell_2$ ball in $\R^p$. \citet{Chandrasekaran}
establish the following theorem linking the Gaussian squared-complexity
of tangent cones and the statistical risk:
%
\begin{theorem}
\label{thmdenoise}
For $\mathbf{x}^\ast\in\bs\subset\R^p$ and with $\bc\subseteq\R
^p$ convex such
that $\bs\subseteq\bc$, we have the error bound
\[
\mathbb{E} \bigl[\bigl\|\mathbf{x}^\ast- \hat{\bx}_n(\bc)
\bigr\|_{\ell
_2}^2 \bigr] \leq\frac{\sigma^2}{n} \G
\bigl(T_\bc\bigl(\mathbf{x}^\ast\bigr) \cap
B_{\ell_2}^p\bigr).
\]
\end{theorem}
This risk bound can be rearranged to yield a way to estimate the number
of data points needed to achieve a given level of risk. In particular,
the theorem implies that if
%
\begin{equation}\label{eqsample-complexity}
n \geq\sigma^2 \G\bigl(T_\bc\bigl(\mathbf{x}^\ast
\bigr) \cap B^p_{\ell_2}\bigr),
\end{equation}
then $\mathbb{E} [\|\mathbf{x}^\ast- \hat{\bx}_n(\bc)\|
_{\ell_2}^2
] \leq1$.
The overall implication is that as the number of data points $n$ grows,
we can back off to computationally cheaper estimators and still control
the statistical risk, simply by choosing the largest $\bc_i$ such
that the right-hand side of equation (\ref{eqsample-complexity}) is
less than $n$.
This yields a time/data tradeoff.

To exemplify the kinds of concrete tradeoffs that can be obtained via
this formalism, \citet{Chandrasekaran} consider a stylized sparse principal
component analysis problem, modeled using the following signal set:
\[
\bs= \bigl\{\Pi M \Pi' \mid \Pi \mbox{ is a } \sqrt{p} \times
\sqrt{p} \mbox{ permutation matrix}\bigr\},
\]
where the top-left $k \times k$ block of $M \in\R^{\sqrt{p} \times
\sqrt{p}}$
has entries equal to $\sqrt{p}/k$ and all other entries are zero.
In Table~\ref{tabsparse-pca} we show the runtimes and sample sizes
associated with
%
\begin{table}
\tablewidth=260pt
\caption{Time-data tradeoffs for the sparse PCA problem, expressed as
a function of the matrix dimension~$p$. See Chandrasekaran and Jordan (\citeyear{Chandrasekaran}) for
details.}
\label{tabsparse-pca}
\begin{tabular*}{\tablewidth}{@{\extracolsep{\fill}}lll@{}}
\hline
$\bc$ & Runtime & $n$ \\
\hline
conv($\bs$) & super-poly($p$) & \mbox{$\sim$}$p^{1/4} \log(p)$ \\[2pt]
Nuclear norm ball & $p^{3/2}$ & \mbox{$\sim$}$p^{1/2}$ \\
\hline
\end{tabular*}
\end{table}
two different convex relaxations of $S$: the convex hull of $S$ and the
nuclear norm ball. The table reveals a time-data tradeoff -- to achieve constant
risk we can either use a more expensive procedure that requires few
data points
or a cheaper procedure that requires few data points or a cheaper procedure
that requires more data points.
%
\begin{table}
\tablewidth=260pt
\caption{Time-data tradeoffs for the cut-matrix denoising problem,
expressed as
a function of the matrix dimension $p$, where $c_1 < c_2 < c_3$.
See \citet{Chandrasekaran} for details.}
\label{tabmatrix-cut}
\begin{tabular*}{\tablewidth}{@{\extracolsep{\fill}}lll@{}}
\hline
$\bc$ & Runtime & $n$ \\
\hline
Cut polytope & super-poly($p$) & $c_1 p^{1/2}$ \\[1pt]
Elliptope & $p^{7/4}$ & $c_2 p^{1/2}$ \\[1pt]
Nuclear norm ball & $p^{3/2}$ & $c_3 p^{1/2}$ \\
\hline
\end{tabular*}
\end{table}
As a second example, consider the cut-matrix denoising problem, where the
signal set is as follows:
\[
\bs= \bigl\{\ba\ba'\mid\ba\in\{-1,+1\}^{\sqrt{p}} \bigr\}.
\]
Table~\ref{tabmatrix-cut} displays the runtimes and sample sizes
associated with
three different convex relaxations of this signal set. Here the
tradeoff is
in the constants associated with the sample size, favoring the cheaper methods.
\citet{Chandrasekaran} also consider other examples, involving variable ordering
and banded covariance matrices. In all of these examples, it seems to
be the
case that the cheaper methods achieve the same risk as more expensive methods
with not very many additional data points.

\section{Discussion}

We have reviewed several lines of research that aim to bring computational
considerations into contact with statistical considerations, with a particular
focus on the matrix-oriented estimation problems that arise frequently
in the
setting of massive data. Let us also mention several other recent theoretical
contributions to the statistics/computation interface. Divide-and-conquer
methodology has been explored by \citet{ChenXie} in the setting of
regression and
classification. Their methods involve estimating parameters on subsets
of data
in parallel and using weighted combination rules to merge these
estimates into
an overall estimates. They are able to show asymptotic equivalence to
an estimator
based on all of the data and also show (empirically) a significant
speed-up via
the divide-and-conquer method. The general idea of algorithmic
weakening via
hierarchies of model families has been explored by several authors;
see, for
example, \citet{Agarwal} and \citet{Shalev-Shwartz}, where the focus is model
selection and classification, and \citet{Amini}, where the focus is
sparse covariance matrix estimation. In all of these lines of work the goal
is to develop theoretical tools that explicitly reveal tradeoffs relating
risk, data and time.

It is important to acknowledge the practical reality that massive datasets
are often complex, heterogeneous and noisy, and the goal of research on
scalability is not that of developing a single methodology that applies
to the analysis of such datasets. Indeed, massive datasets will require
the full range of statistical methodology to be brought to bear in order
for assertions of knowledge on the basis of massive data analysis to be
believable.
The problem is that of recognizing that the analysts of massive data will
often be interested not only in statistical risk, but in risk/time tradeoffs,
and that the discovery and management of such tradeoffs can only be achieved
if the algorithmic character of statistical methodology is fully acknowledged
at the level of the foundational principles of the field.

\section*{Acknowledgements}

I wish to acknowledge numerous colleagues who have helped to shape the
perspective presented here, in particular Venkat Chandrasekaran, Ariel Kleiner,
Lester Mackey, Purna Sarkar and Ameet Talwalkar.



\printhistory

\end{document}